\PassOptionsToPackage{authoryear,longnamesfirst}{natbib}
\let\savedBibliographyStyle\bibliographystyle
\renewcommand{\bibliographystyle}[1]{}
\documentclass[10pt,twoside]{predoc2}
\let\bibliographystyle\savedBibliographyStyle
\hypersetup{allcolors=black}
\usepackage{iftex}
\ifXeTeX
\fi
\usepackage{xurl}
\usepackage{graphicx}
\usepackage{amsmath}
\usepackage{subcaption}
\usepackage{xcolor}
\usepackage{placeins}
\usepackage{wrapfig}

\usepackage{etoolbox}
\ifXeTeX
\usepackage{noindentafter}
\forcsvlist{\NoIndentAfterEnv}{figure,figure*,table,table*,equation,equation*,align,align*,gather,gather*,multline,multline*,displaymath}
\NoIndentAfterCmd{\FloatBarrier}
\makeatletter
\pretocmd{\@afterheading}{\@afterindentfalse}{}{}
\makeatother
\else
\providecommand{\phantomsubcaption}{}
\fi

\usepackage[percent]{overpic}

\newcounter{panel}

\newcommand{\runauthor}{Reznik et al.}
\newcommand{\runtitle}{Adversarial robustness does not entail human-aligned categorization}
\title{Adversarial robustness does not entail human-aligned categorization}
\newcommand{\itayAffiliation}{Institute for Applied AI Research\\
  \textit{Ben-Gurion University of the Negev}}
\newcommand{\razAffiliation}{\textit{Intuit}}
\newcommand{\mosheAffiliation}{Institute for Applied AI Research\\
  Data Science Research Center\\
  \textit{Ben-Gurion University of the Negev}}
\newcommand{\talAffiliation}{Department of Industrial Engineering and Management\\
  School of Brain Sciences and Cognition\\
  Data Science Research Center\\
  \textit{Ben-Gurion University of the Negev}}
\ifXeTeX
\newcommand{\manuscriptAuthorBlock}[3]{%
  \begin{minipage}[t]{0.48\textwidth}
    \raggedright
    {\normalsize\sffamily\bfseries #1\par}
    \vspace{2pt}
    {\small\rmfamily #2\par}
    \vspace{2pt}
    {\small\rmfamily #3\par}
  \end{minipage}%
}
\makeatletter
\renewcommand{\@author}{%
  \manuscriptAuthorBlock
    {\href{https://orcid.org/0009-0002-9375-8795}{Itay Reznik}}
    {\itayAffiliation}
    {\href{mailto:reznikit@post.bgu.ac.il}{reznikit@post.bgu.ac.il}}%
  \hfill
  \manuscriptAuthorBlock
    {\href{https://orcid.org/0000-0002-4818-9338}{Raz Lapid}}
    {\razAffiliation}
    {\href{mailto:raz_lapid@intuit.com}{raz\_lapid@intuit.com}}%
  \par\vspace{12pt}\noindent
  \manuscriptAuthorBlock
    {\href{https://orcid.org/0000-0003-1811-472X}{Moshe Sipper}}
    {\mosheAffiliation}
    {\href{mailto:sipper@bgu.ac.il}{sipper@bgu.ac.il}}%
  \hfill
  \manuscriptAuthorBlock
    {\href{https://orcid.org/0000-0002-7940-7473}{Tal Golan}}
    {\talAffiliation}
    {\href{mailto:golan.neuro@bgu.ac.il}{golan.neuro@bgu.ac.il}\par
     Corresponding author}%
}
\makeatother
\else
\author[itay-affiliation,itay-email]{\textbf{\href{https://orcid.org/0009-0002-9375-8795}{Itay Reznik}}}
\affil[itay-affiliation]{\itayAffiliation}
\affil[itay-email]{\href{mailto:reznikit@post.bgu.ac.il}{reznikit@post.bgu.ac.il}}
\author[raz-affiliation,raz-email]{\textbf{\href{https://orcid.org/0000-0002-4818-9338}{Raz Lapid}}}
\affil[raz-affiliation]{\razAffiliation}
\affil[raz-email]{\href{mailto:raz_lapid@intuit.com}{raz\_lapid@intuit.com}}
\author[moshe-affiliation,moshe-email]{\textbf{\href{https://orcid.org/0000-0003-1811-472X}{Moshe Sipper}}}
\affil[moshe-affiliation]{\mosheAffiliation}
\affil[moshe-email]{\href{mailto:sipper@bgu.ac.il}{sipper@bgu.ac.il}}
\author[tal-affiliation,tal-correspondence]{\textbf{\href{https://orcid.org/0000-0002-7940-7473}{Tal Golan}}}
\affil[tal-affiliation]{\talAffiliation}
\affil[tal-correspondence]{\href{mailto:golan.neuro@bgu.ac.il}{golan.neuro@bgu.ac.il}\\Corresponding author}
\fi
\date{}

\begin{document}
\maketitle
\thispagestyle{firstpage}
\hypersetup{pdfauthor={Itay Reznik; Raz Lapid; Moshe Sipper; Tal Golan},pdfsubject={}}

\begin{abstract}
Adversarial training can improve some forms of correspondence between artificial and biological vision, but whether robust classifiers thereby acquire human-aligned classification behavior remains unclear. We generated 1,008 controversial stimuli by optimizing latent inputs to a pretrained GAN to elicit high-confidence classification disagreements between pairs of robust ImageNet classifiers. We compared four high-performing adversarially robust ImageNet classifiers---ResNet-101, ViT-B, ConvNeXt-L, and Swin-L---with two-alternative category judgments from 60 participants. We first evaluated each classifier on stimuli optimized to elicit disagreement between that classifier and another model. Across the four classifiers, accuracy in predicting individual participants' single-trial category choices was approximately at chance (48.6--51.3\%), whereas majority judgments from other participants predicted those responses with 85.0--86.8\% accuracy. An analysis incorporating model prediction strength and participants' reported confidence found weak positive model--human correspondence for the two convolutional models. On stimuli optimized to elicit disagreement between pairs of other models, all four robust classifiers predicted human responses above chance but less accurately than predictions based on other participants' majority judgments; the same pattern held for 17 additional RobustBench models not involved in stimulus generation. Across the 17 additional models, mean model--participant agreement was 67.3\%, versus 81.3\% mean participant--participant agreement on the same images. Together, the results identify model-specific failures of human alignment and a model--human mismatch that persists across robust classifiers and substantially exceeds disagreement among human observers. These findings show that adversarial robustness alone does not guarantee human-aligned categorization and suggest that robustness and human alignment should be evaluated as distinct properties of visual recognition systems.

\end{abstract}

\begin{center}
\includegraphics[width=.65\textwidth]{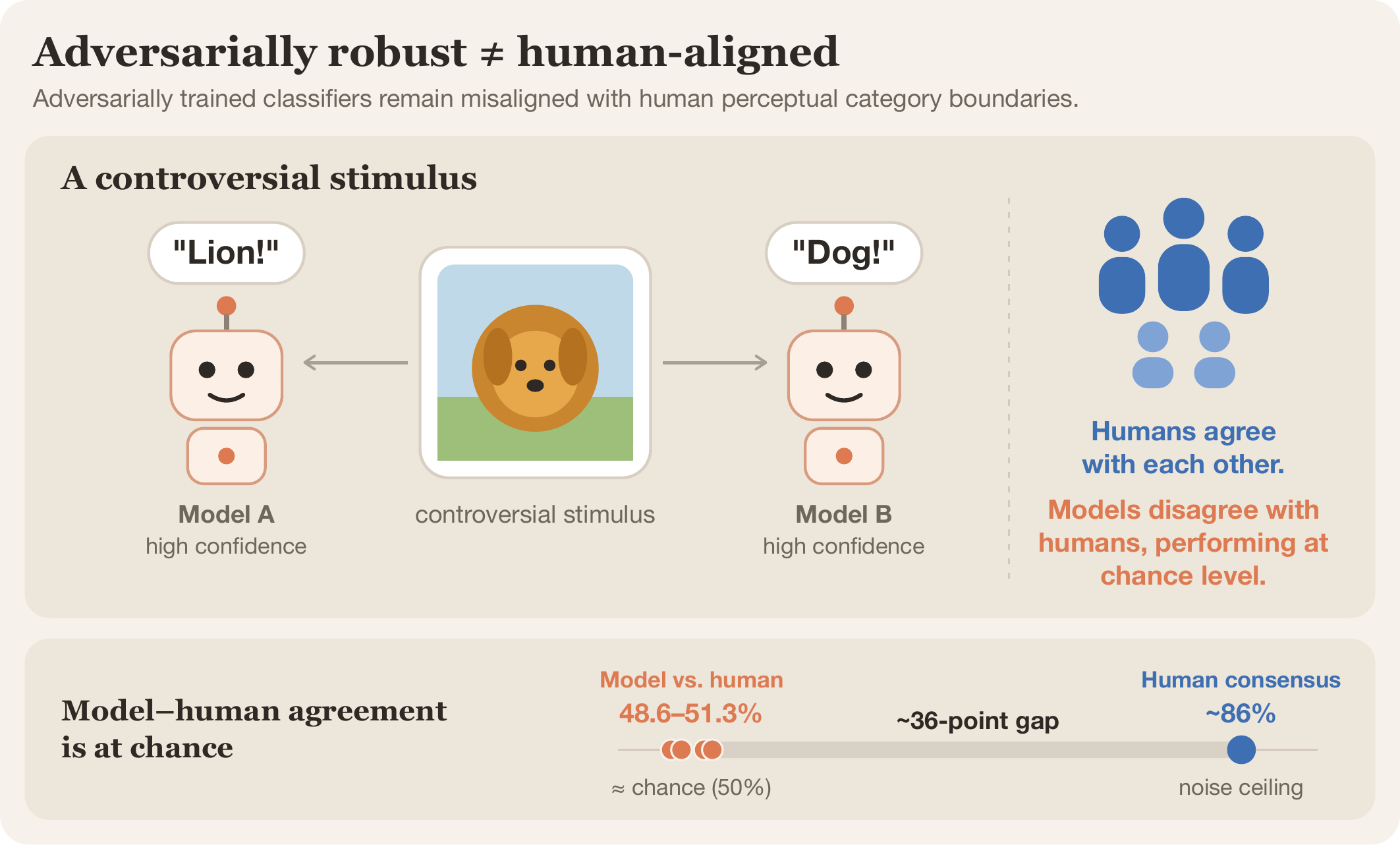}
\end{center}

\section{Introduction}
\label{sec:introduction}

Deep neural network classifiers have reached or exceeded human performance on benchmark object-recognition tasks such as ImageNet \citep{russakovsky2015imagenet,shankar2020evaluating,geirhos2021partial,kul2026comparison}, yet adversarial examples reveal that model and human classification behavior can nevertheless differ sharply: small perturbations can change a model's classification while leaving the image perceptually unchanged to human observers \citep{szegedy2014intriguing,goodfellow2015explaining}. Adversarial training addresses this vulnerability in a narrowly defined sense, making model classifications stable under norm-bounded perturbations to natural images (\citealp{madry2018towards}; see critique by \citealp{tramer2020fundamental}). But does the robustness resulting from adversarial training imply that model and human categorization boundaries are aligned?

Several findings suggest that adversarially robust networks are indeed more closely aligned with biological vision. Images optimized to control the internal representations or outputs of adversarially trained networks produce effects that generalize to human behavior \citep{tsipras2019robustness,feather2023model,gaziv2023strong} and to non-human primate cortical responses \citep{guo2022adversarially,gaziv2025noninvasive}, with recent work specifically showing stronger neural control with adversarially trained than with non-robust models \citep{prince2026parametric}. However, such transfer can also arise from partial correspondence between the systems and therefore does not establish that their categorization boundaries are aligned. Whether robust models and humans share the same categorization boundaries therefore remains unresolved.

Here, we use \emph{controversial stimuli} to test more severely whether current adversarially trained models are aligned with human visual categorization behavior. Controversial stimuli are inputs specifically optimized to expose disagreements between models or between models and humans \citep{golan2020controversial,golan2026making}. \citet{golan2020controversial} applied this approach to models trained on relatively simple, low-resolution datasets---MNIST \citep{lecun1998gradient} and CIFAR-10 \citep{krizhevsky2009learning}---and generated images using gradient descent directly in pixel space. When non-robust models were contrasted with adversarially trained models, humans tended to classify the resulting stimuli in agreement with the robust model. However, when robust models were pitted against one another, the resulting images were often not recognizable to humans, a case analogous to ``fooling images'', in which an unrecognizable image elicits a specific high-confidence classification from a model \citep{nguyen2015deep}. Although confident classification of an image containing no human-recognizable object reveals a clear discrepancy between model and human judgments, it provides a less direct test of categorization-boundary alignment: discriminative classifiers are trained to distinguish among object classes, not necessarily between those classes and arbitrary non-objects.

In our work, we generate controversial stimuli under a natural-image prior by optimizing the latent input to a pre-trained class-conditional BigGAN \citep{brock2019large} rather than optimizing directly in pixel space. The resulting images do not necessarily look natural, but humans nevertheless classify them consistently. Related work has demonstrated that optimizing within a pretrained GAN's latent space enables naturalistic, gradient-free black-box adversarial attacks on object detectors \citep{lapid2026patch}. Here, we adopt a generative-prior approach for a different purpose: exposing category disagreements between robust classifiers and evaluating their correspondence with human judgments. We use the Covariance Matrix Adaptation Evolution Strategy (CMA-ES) \citep{hansen2001completely}, a gradient-free, population-based evolutionary optimizer. Because CMA-ES requires only objective-function evaluations rather than gradients, both the generator and classifiers can remain black boxes. Gradient-free evolutionary optimization has previously been shown to enable query-efficient black-box adversarial attacks on image classifiers without requiring access to model gradients \citep{lapid2022evolutionary}. Consequently, the procedure is unaffected by potential gradient obfuscation \citep{athalye2018obfuscated} and can also be applied to classifiers available only through inference APIs, provided their relevant outputs (e.g., logits or class probabilities) are accessible. The population-based search in CMA-ES evaluates multiple candidate solutions at each generation, reducing sensitivity to specific starting conditions.

We test a representative set of four state-of-the-art adversarially robust ImageNet classifiers, selected for high adversarial accuracy and including two Convolutional Neural Networks (CNNs) and two Transformers. Across all six pairwise model combinations, we optimize images that elicit sharply different category predictions and then ask human observers to classify the same images. Humans largely agree with one another on the classification of these stimuli, yet individual models agree with human judgments mostly at chance. Thus, the categorization boundaries of current state-of-the-art adversarially robust models remain markedly misaligned with those of humans.

Our main contributions are:
\begin{enumerate}
\item \textbf{Gradient-free, prior-informed controversial-stimulus optimization.} We extend controversial-stimulus optimization to ImageNet-scale robust classifiers by searching under a natural-image generative prior and using derivative-free optimization that does not require differentiable models or access to model gradients.

\item \textbf{Direct behavioral evidence for misaligned model--human categorization boundaries.} We show that current state-of-the-art adversarially robust models remain markedly misaligned with human categorization boundaries: humans classify the optimized stimuli consistently, whereas individual models agree with human judgments mostly at chance.

\item \textbf{New human benchmark data on controversial stimuli.} We collect 10,080 human judgments on 1,008 controversial stimuli, providing data for evaluating model--human alignment in future vision models.

\end{enumerate}

\section{Methods}
\label{sec:methodology}

\begin{figure}[t]
\centering
\includegraphics[width=0.9\linewidth, trim=0 0.5cm 0 0.5cm, clip]{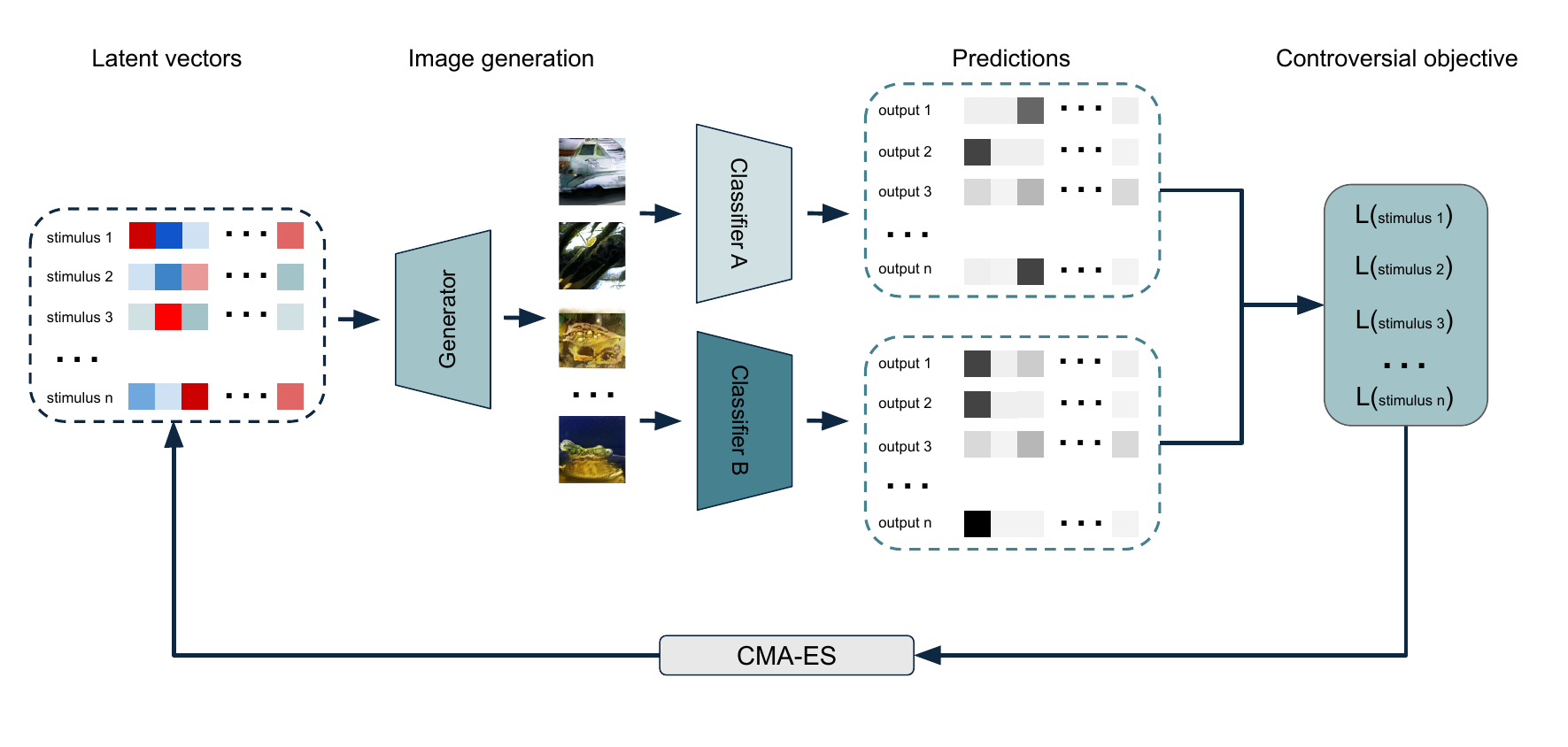}
\caption{\textbf{Controversial-stimulus generation.} To synthesize a stimulus, two classifiers were assigned different target classes. CMA-ES sampled candidate inputs to a fixed, pre-trained BigGAN generator, and each generated image was evaluated by both classifiers. The objective favored images for which each classifier assigned high probability to its assigned target class. The resulting objective values were returned to CMA-ES, which iteratively updated its sampling distribution to search the generator input space for increasingly controversial stimuli. The resulting stimuli were then used to test how the conflicting model predictions aligned with human visual categorization.}
\label{fig:optflow}
\end{figure}
\FloatBarrier

\subsection{Robust Classification Models}

We selected four high-performing adversarially trained ImageNet classifiers: two CNNs---ResNet-101 \citep{he2016deep} and ConvNeXt-L \citep{liu2022convnet}---and two Transformer-based models---ViT-B/16 \citep{dosovitskiy2021image} and Swin-L \citep{liu2021swin}.

We obtained the ResNet-101 and ViT-B weights directly from ARES \citep{liu2025comprehensive} and loaded ConvNeXt-L and Swin-L through RobustBench \citep{croce2021robustbench}.

All four models were adversarially trained on ImageNet under an $\ell_\infty$ threat model with perturbation budget $\epsilon=4/255$ \citep{croce2021robustbench,liu2025comprehensive}. Training used three-step projected gradient descent (PGD) with the ARES default step size. The models shared the threat model and evaluation protocol, although their training recipes were not identical: ResNet-101 used a shorter, SGD-based training schedule, whereas ViT-B/16, ConvNeXt-L, and Swin-L used 300-epoch AdamW training with cosine scheduling and stronger augmentation. Table~\ref{tab:robust-models} summarizes their clean and adversarial accuracies. The models were trained on $224\times224$ inputs normalized with the ImageNet channel means and standard deviations. We used the same input format during both stimulus synthesis and model evaluation.

\begin{table}[!htbp]
\centering
\caption{Robust ImageNet classifiers used in the study. Parameters are reported in millions. Clean accuracy is top-1 accuracy on unperturbed images. PGD100 is top-1 accuracy under a 100-step projected gradient descent attack, and AutoAttack is top-1 accuracy under AutoAttack \citep{croce2020reliable}. Both adversarial evaluations used an $\ell_\infty$ perturbation budget of $\epsilon=4/255$. All three accuracy values are the ARES-reported results on the same random sample of 1,000 images drawn without replacement from the ImageNet validation set; we independently validated all three results on that image set.}
\label{tab:robust-models}
\small
\setlength{\tabcolsep}{6pt}
\renewcommand{\arraystretch}{1.15}
\begin{tabular}{l r r r r}
\hline
Model & Parameters (M) & Clean acc. (\%) & PGD100 acc. (\%) & AutoAttack acc. (\%) \tabularnewline
\hline
ResNet-101 & 44.5 & 71.0 & 46.5 & 42.2 \tabularnewline
ViT-B/16 & 86.6 & 74.7 & 52.2 & 49.7 \tabularnewline
ConvNeXt-L & 197.8 & 78.8 & 61.7 & 60.1 \tabularnewline
Swin-L & 196.5 & 79.7 & 63.9 & 62.3 \tabularnewline
\hline
\end{tabular}
\end{table}

We also evaluated 17 additional adversarially robust ImageNet classifiers distributed through RobustBench \citep{croce2021robustbench}. These models were evaluated on the final stimulus set but were not used to generate or select the stimuli. Of the four models used for stimulus generation, only ConvNeXt-L and Swin-L appeared on the RobustBench leaderboard. We considered the 24 highest-ranked ImageNet $\ell_\infty$ models on the leaderboard (accessed in February 2025), ranked by AutoAttack robust accuracy at $\epsilon=4/255$, and excluded the seven that were not independent of the generating models: ConvNeXt-L and Swin-L, their smaller ConvNeXt-B and Swin-B variants, and three models built on the ConvNeXt-L or Swin-L checkpoints. The remaining 17 models are listed in Supplementary Table~\ref{tab:additional-models}.

\subsection{Stimulus Generation}

For each controversial stimulus, we selected two classifiers and assigned a different target class to each classifier. We then optimized the input to a fixed class-conditional BigGAN to generate an image for which each classifier assigned high probability to its target class (Fig.~\ref{fig:optflow}). The optimization therefore searched within the generator input space rather than directly in pixel space. We next describe generator and classifier preprocessing, generator-input parameterization, the controversial-stimulus objective, CMA-ES optimization, and the optimization runs used for image selection.

\paragraph{Generator and classifier preprocessing.}
We used BigGAN \citep{brock2019large}, specifically the \texttt{biggan-deep-128} checkpoint loaded through the PyTorch Pretrained GANs package \citep{melas2022finding}. The model was pretrained on ImageNet-1K. It takes a 128-dimensional noise vector and a 1,000-dimensional class-conditioning vector as inputs and produces $128\times128$ RGB images. Before being passed to the classifiers, generator outputs were mapped from $[-1,1]$ to $[0,1]$, resized to $256\times256$ pixels using bicubic interpolation, center-cropped to $224\times224$, and normalized as required by the corresponding classifier implementation.

\paragraph{Generator-input parameterization.}
CMA-ES optimized a 130-dimensional search vector. Its first 128 dimensions parameterized the BigGAN noise input, and its remaining two dimensions parameterized the 1,000-dimensional class-conditioning vector:
\begin{equation}
\label{eq:search-vector}
\mathbf{u}=[\boldsymbol{\epsilon};\mathbf{a}],
\end{equation}
where $\boldsymbol{\epsilon}\in\mathbb{R}^{128}$ contained the noise parameters and $\mathbf{a}=(a_A,a_B)\in\mathbb{R}^{2}$ contained the two class-conditioning scores. Before passing the noise input to BigGAN, we projected $\boldsymbol{\epsilon}$ onto the closed $\ell_2$ ball of radius $R=11.5$ to obtain $\tilde{\boldsymbol{\epsilon}}$:
\begin{equation}
\label{eq:noise-projection}
\tilde{\boldsymbol{\epsilon}}=
\begin{cases}
\boldsymbol{\epsilon}, & \lVert\boldsymbol{\epsilon}\rVert_2\leq R,\\
R\,\boldsymbol{\epsilon}/\lVert\boldsymbol{\epsilon}\rVert_2,
& \lVert\boldsymbol{\epsilon}\rVert_2>R.
\end{cases}
\end{equation}
BigGAN was trained with noise drawn from a standard normal prior. For a 128-dimensional vector drawn from this prior, $\mathbb{E}\lVert\boldsymbol{\epsilon}\rVert_2\approx\sqrt{127.5}\approx11.3$. We therefore set $R=11.5$, slightly above this expected norm; vectors with norms greater than $R$ were radially rescaled to norm $R$. CMA-ES itself operated on the raw, unconstrained vector $\mathbf{u}$.
We passed the two class-conditioning scores through a softmax with temperature $T=3.0$:
\begin{equation}
\label{eq:conditioning-softmax}
\tilde{\mathbf{a}}=\operatorname{softmax}(\mathbf{a}/T).
\end{equation}
We then constructed the class-conditioning vector $\mathbf{y}$ by setting $y_A=\tilde{a}_A$ and $y_B=\tilde{a}_B$, with all other entries set to zero. The generated image was
\begin{equation}
\label{eq:generator-output}
\mathbf{x}=G_\psi(\tilde{\boldsymbol{\epsilon}},\mathbf{y}).
\end{equation}

\paragraph{Controversial-stimulus objective.}
Let $f_\theta$ and $f_\phi$ denote the two selected classifiers, and let $A$ and $B$ denote their respective, distinct target classes. The quantities $f_\theta(A\mid\mathbf{x})$ and $f_\phi(B\mid\mathbf{x})$ are the probabilities assigned to these target classes by the corresponding full 1,000-class softmax distributions. For a candidate vector $\mathbf{u}$ and its generated image $\mathbf{x}(\mathbf{u})$, we minimized
\begin{equation}
\label{eq:loss_func}
\mathcal{L}(\mathbf{u})
= -\log f_\theta(A\mid\mathbf{x}(\mathbf{u}))
  -\log f_\phi(B\mid\mathbf{x}(\mathbf{u})).
\end{equation}
Minimizing this objective is equivalent to maximizing
$\log\!\left[f_\theta(A\mid\mathbf{x}(\mathbf{u}))f_\phi(B\mid\mathbf{x}(\mathbf{u}))\right]$.
Because neither probability can exceed one, this quantity approaches zero from below only when both target-class probabilities approach one---that is, when each classifier assigns high probability to its assigned target class.

\paragraph{CMA-ES optimization.}
We optimized $\mathbf{u}$ using version 1.5.0 of the \texttt{cma-es} Python package \citep{strock2021cmaes}. The initial mean of the 130-dimensional sampling distribution comprised a 128-dimensional noise sample drawn from $\mathcal{N}(\mathbf{0},I)$ and conditioning scores $(0.5,0.5)$. The initial step size was 1.0. Population size was set according to the package's default rule, $\lambda=\lfloor3\ln(D)+8\rfloor$, which yielded 22 candidates per generation for $D=130$. Optimization terminated when one of the package's default stopping criteria was met or after 500 generations, whichever occurred first.

The final image for each run was generated by applying the same generator-input transformations to the mean of the CMA-ES sampling distribution at termination.

\paragraph{Optimization runs and image selection.}
The four models formed six unordered model pairs. For each model pair, we considered all ordered pairs of distinct target classes drawn from eight ImageNet classes---\emph{indigo bird}, \emph{tailed frog}, \emph{golden retriever}, \emph{lion}, \emph{container ship}, \emph{garbage truck}, \emph{race car}, and \emph{military plane}---giving 56 ordered class pairs. We performed eight optimization runs from different random initializations for every model-pair--class-pair combination, for a total of $6\times56\times8=2{,}688$ runs. An image was eligible for selection if $f_\theta(A\mid\mathbf{x})>f_\theta(B\mid\mathbf{x})$ and $f_\phi(B\mid\mathbf{x})>f_\phi(A\mid\mathbf{x})$. Thus, for each classifier, the probability of its target class had to exceed that of the other target class. Eligible images were ranked by the controversiality score
\begin{equation}
\label{eq:controversiality-score}
C(\mathbf{x})=\min\{f_\theta(A\mid\mathbf{x}),f_\phi(B\mid\mathbf{x})\}.
\end{equation}
For each combination, we retained the three highest-scoring images. The eight runs allowed us to select 991 of the planned 1,008 images; some combinations had fewer than three eligible images. We performed additional runs for those combinations until three eligible images could be selected for each. This yielded $6\times56\times3=1{,}008$ experimental stimuli.

\subsection{Behavioral Experiment}
\label{sec:Experimentation}

To evaluate model--human alignment, we collected category judgments and confidence ratings for the 1,008 experimental stimuli in an online forced-choice categorization experiment with two candidate categories per image.

\paragraph{Participants.}
Participants were recruited through Prolific. To participate, individuals were required to be 18--50 years old, reside in the United States, have English as a first language, and report normal or corrected-to-normal vision and no color-vision deficiency. They were also required to have a Prolific approval rate of 90--100\% and at least 10 previous submissions. Sixty-four participants took part in the study. After applying the quality-control procedure described below, the analyzed sample included 60 participants (age range 21--50 years; mean age 37.05 years; 30 female). Participants were compensated at a rate of \textdollar 12 per hour; sessions lasted approximately 20 minutes. The Ben-Gurion University of the Negev Human Subjects Research Committee approved the study, and all participants provided informed consent.

\paragraph{Stimulus assignment and procedure.}
When a participant began the experiment, the online assignment procedure selected 168 target stimuli from the full set of 1,008, prioritizing those that had been presented to the fewest participants and breaking ties at random. In the final dataset, after exclusion of participants with low-effort responses, each stimulus was judged by exactly 10 participants.

Each session began with task instructions delivered through an instructional video. Participants then completed five practice trials, which were excluded from all analyses. The main experiment consisted of 203 trials: 168 target trials, five attention-check trials randomly interspersed throughout the session, and 30 repeat trials presented at the end. Only responses from the 168 target trials were included in the model--human alignment analyses. On each trial, an image was presented centrally on the screen alongside two category labels (Supplementary Fig.~\ref{fig:behavioral-trial}). For target and repeat trials, the category labels corresponded to the two target classes used to generate the image. Participants reported both their category judgment and confidence by clicking one of six buttons: each category label was paired with three confidence options---\emph{certain}, \emph{hesitating}, and \emph{guessing}. The left-right positions of the two labels were randomized across trials, and target-trial order was randomized within each session. Attention checks used unambiguous natural images for which one of the two labels was correct. The repeat stimuli were randomly sampled from the participant's target trials and were used to assess within-participant consistency. Trials were self-paced, with no time limit. The experiment was implemented in PsychoPy and administered online through Pavlovia.

\paragraph{Quality control.}
Before collecting the main cohort, we calibrated an automated quality-control procedure in pilot testing. The procedure combined four indicators: responses faster than one second, errors on the five attention-check trials, inconsistency between responses to the original and repeated presentations, and runs of more than five identical responses. The calibrated procedure was then applied unchanged to the main cohort, excluding 4 of 64 participants and leaving 60 for analysis.

\subsection{Quantifying Model--Human Alignment}
\label{sec:quantifying-alignment}

We quantified model--human alignment using two complementary measures. Mean binary model--human agreement considered only whether the model prediction matched the participant's category choice, whereas signed-rank cosine similarity (SRCS) also incorporated human and model confidence levels. For both measures, we estimated lower and upper bounds on the human noise ceiling (i.e., a reliability bound on predictive validity) and quantified uncertainty using a two-factor bootstrap.

\paragraph{Mean binary agreement.}
For each model and stimulus, the predicted category was the candidate class with the larger logit. Trial-level agreement was scored as 1.0 when the predicted category matched the participant's category choice and 0 otherwise. Mean binary agreement was computed by averaging these scores across trials within each participant and then across participants. It ranges from 0 to 1: a value of 1 indicates a match on every trial, 0.5 is chance for a two-category choice, and 0 indicates a mismatch on every trial.

\paragraph{Signed-rank cosine similarity.}
Following \citet{golan2023testing}, we constructed a human-response vector $\mathbf{y}^{(h)}$ and a model category-preference vector $\mathbf{y}^{(m)}$ over the trials completed by each participant. In both vectors, the sign encoded category identity, and the magnitude encoded confidence. For each trial, one candidate category was assigned a positive sign and the other a negative sign. This assignment does not affect the resulting SRCS values, provided it is the same in both vectors. Entries in the human-response vector, $\mathbf{y}^{(h)}$, were $-1$, $-2$, or $-3$ for choices of the negative category and $+1$, $+2$, or $+3$ for choices of the positive category; magnitudes 1, 2, and 3 represented \emph{guessing}, \emph{hesitating}, and \emph{certain} responses, respectively. The corresponding entry in the model category-preference vector, $\mathbf{y}^{(m)}$, was the logit of the positive category minus the logit of the negative category. This difference equals the log probability ratio between the two candidate categories. The average signed-rank vectors were
\begin{equation}
\label{eq:signed-ranks}
\begin{aligned}
\bar{\mathbf{r}}^{(h)}
&=\operatorname{sign}\!\left(\mathbf{y}^{(h)}\right)
\odot\operatorname{rank}_{\mathrm{avg}}\!\left(\left|\mathbf{y}^{(h)}\right|\right),\\
\bar{\mathbf{r}}^{(m)}
&=\operatorname{sign}\!\left(\mathbf{y}^{(m)}\right)
\odot\operatorname{rank}_{\mathrm{avg}}\!\left(\left|\mathbf{y}^{(m)}\right|\right).
\end{aligned}
\end{equation}
Here, $\operatorname{rank}_{\mathrm{avg}}$ denotes the rank transformation with average ranks assigned to tied magnitudes, and $\odot$ denotes elementwise multiplication. For a participant who completed $n$ trials, SRCS was
\begin{equation}
\label{eq:srcs}
\operatorname{SRCS}(m,h)
=\frac{\bar{\mathbf{r}}^{(m)}\cdot\bar{\mathbf{r}}^{(h)}}
{\displaystyle\sum_{k=1}^{n}k^2}.
\end{equation}
The denominator is the squared norm of the complete rank sequence $1,\ldots,n$, rather than the squared norm of the average-rank vector; this normalization yields the expected SRCS under random tie-breaking \citep{golan2023testing}. SRCS ranges from $-1$ to $1$: larger positive values indicate closer correspondence between the signed rankings, values near zero indicate no systematic correspondence, and negative values indicate opposing signed-rank patterns. SRCS was computed for each participant and then averaged across participants.

\paragraph{Human noise ceiling.}
To establish bounds on achievable model performance given the inherent inter-subject variability of human perceptual judgments, we estimated the human noise ceiling. We estimated two bounds by predicting each participant's response on each trial from the majority response to the same stimulus. For the lower bound, the prediction was the majority vote of the other nine participants who evaluated that stimulus. For the upper bound, the prediction was the majority vote of all 10 participants, including the participant whose response was being evaluated.
For each bound, trial-level agreement was calculated with the human majority-vote prediction in place of the model prediction. A tied majority received an agreement score of 0.5. As in the calculation of mean binary model--human agreement, scores were first averaged across trials within each participant and then across participants.

Lower and upper bounds on the SRCS noise ceiling were estimated analogously using normalized signed-rank response patterns, following \citet{golan2023testing}. For each participant's trial set, the lower bound used the average response pattern of the other raters of each stimulus, whereas the upper bound used the average response pattern including that participant.

\paragraph{Statistical inference.}
Inferential uncertainty was estimated using a two-factor percentile bootstrap with 10,000 resamples. Participants were resampled from the full participant pool, and stimuli were resampled separately within each of the six groups defined by the pair of models used to generate them. The same resamples were used for all models, with participant scores and noise-ceiling bounds recomputed for each resample. We report 95\% confidence intervals and finite-sample two-sided bootstrap tests of mean binary agreement against chance (0.5), SRCS against 0, each measure against the lower bound on its noise ceiling, and pairwise differences between models. For each alignment measure, we adjusted $p$-values using the Benjamini--Hochberg procedure \citep{benjamini1995controlling} to control the false-discovery rate (FDR) at .05, treating comparisons with chance, the lower noise-ceiling bound, and other models as separate families within each analysis.

\section{Results}
\label{sec:results}

\subsection{Controversial-stimulus optimization elicited opposing model predictions}
We generated three experimental stimuli for each combination of the six model pairs and 56 ordered target-class pairs. All 1,008 stimuli elicited the intended disagreement: for each of the two generating models, the probability assigned to its target class exceeded that assigned to the alternative target class (Supplementary Fig.~\ref{fig:pairwise-target-probabilities}).

\begin{figure}[p]
    \centering
    \includegraphics[width=\linewidth]{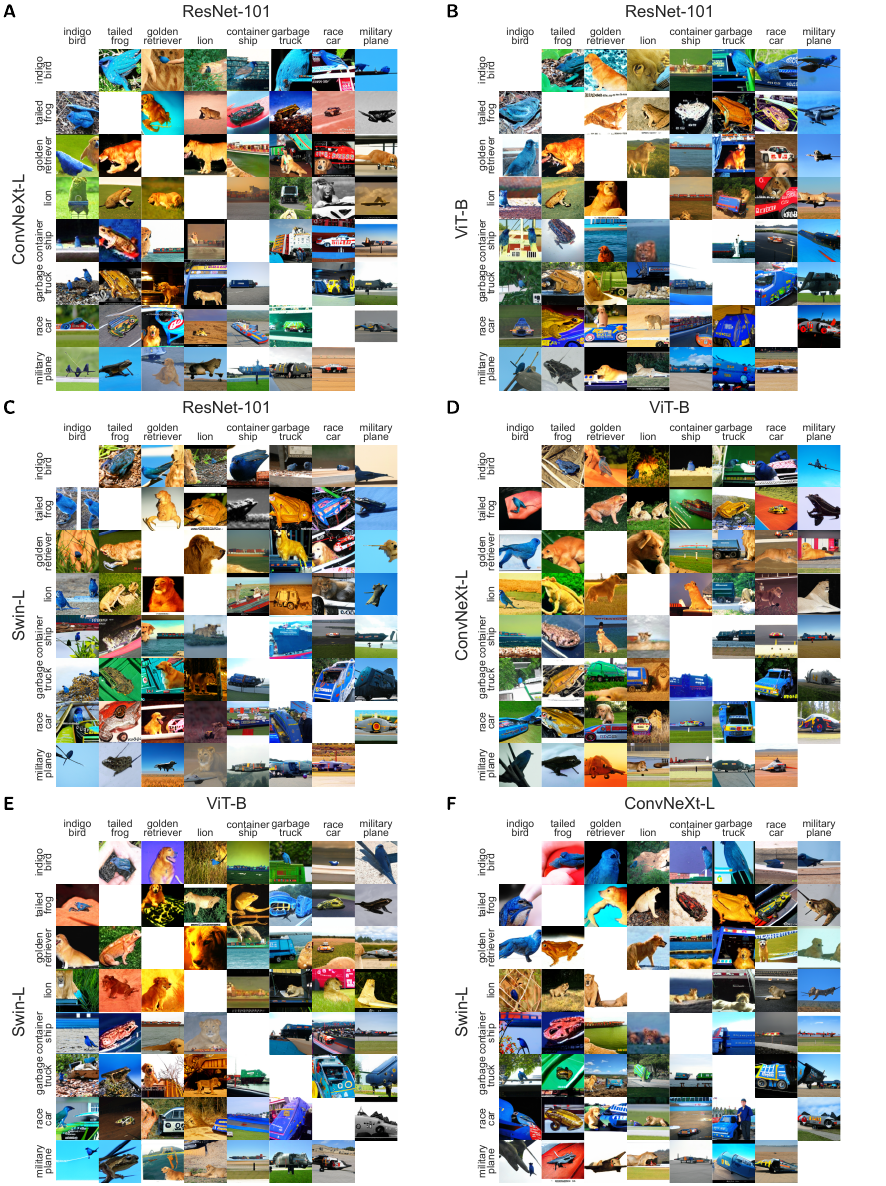}
    \caption{\textbf{Generated controversial stimuli for all six model pairs.} Each panel corresponds to one model pair; rows and columns denote the classes favored by the models on the vertical and horizontal axes, respectively. Diagonal cells are omitted because the two target classes always differed. Across the displayed stimuli, models assigned a mean probability of 0.99 to their target classes after normalization over the two candidate classes. Each cell shows the most controversial (Eq.~\ref{eq:controversiality-score}) of three stimuli used in the behavioral experiment for that model pair and ordered target-class pair.}
    \label{fig:all_comparisons_stimulus_montage}
\end{figure}

The GAN-generated stimuli all contained object-like forms (Fig.~\ref{fig:all_comparisons_stimulus_montage}). Informal visual inspection suggests that some forms are readily recognizable, whereas others are ambiguous or difficult to describe. To establish the distribution of human category judgments for each stimulus, we conducted an online behavioral experiment with 60 participants.

\subsection{Current robust models are poorly aligned with human category judgments}
\begin{figure}[ht!]
    \centering
    \renewcommand{\thesubfigure}{\Alph{subfigure}}
    \begin{subfigure}[t]{0.5\linewidth}
        \centering
        \makebox[\linewidth][l]{\sffamily\bfseries\small A}
        \includegraphics[width=\linewidth]{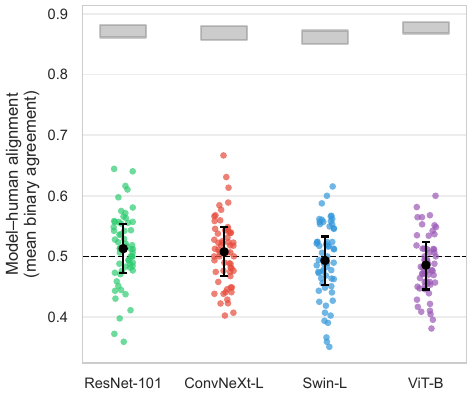}
        \phantomsubcaption\label{fig:model_accuracies}
    \end{subfigure}
    \hfill
    \begin{subfigure}[t]{0.49\linewidth}
        \centering
        \makebox[\linewidth][l]{\sffamily\bfseries\small B}
        \includegraphics[width=\linewidth]{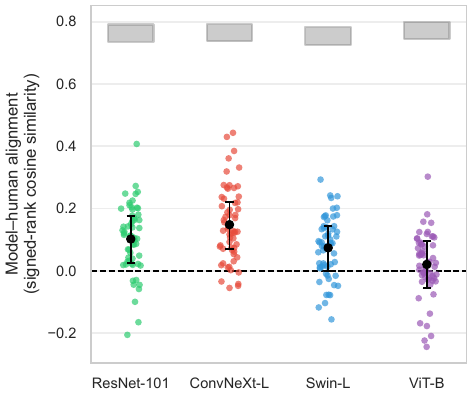}
        \phantomsubcaption\label{fig:signed-participating}
    \end{subfigure}

    \caption{\textbf{Model--human alignment for the four models.} Both panels evaluate each model only on stimuli generated to target it. \textbf{(A)} Mean binary model--human agreement. \textbf{(B)} Signed-rank cosine similarity (SRCS). Colored points show participant-level values; black markers and error bars show group means and 95\% two-factor bootstrap confidence intervals. The dashed line marks chance-level performance. Gray bars span the lower and upper bounds on the human noise ceiling.}
    \label{fig:model_comparison}
\end{figure}

Each stimulus was judged by exactly 10 participants. We first evaluated each model on the stimuli generated to target it. Mean binary model--human agreement was near chance for every model: 51.3\% for ResNet-101 (95\% CI [47.3, 55.3]), 50.8\% for ConvNeXt-L (95\% CI [46.7, 54.8]), 49.3\% for Swin-L (95\% CI [45.3, 53.3]), and 48.6\% for ViT-B (95\% CI [44.5, 52.4]; Fig.~\ref{fig:model_comparison}A). Mean model--human agreement did not differ significantly from chance for any of the four models (all uncorrected $p\geq.453$). None of the six pairwise model comparisons was significant after FDR correction (minimum adjusted $p=.871$).

Could the near-chance binary agreement simply reflect inconsistent human responses? Inter-participant disagreement about the perceived category, or even guessing by uncertain participants, would limit the predictability of human responses. However, single-trial human judgments were highly predictable from the judgments of other participants who viewed the same stimulus. The lower human noise-ceiling bounds ranged from 85.0\% to 86.8\%, and all four model estimates were below their corresponding lower bounds (all adjusted $p=.0002$). For each generating-model pair, we asked whether human responses matched one model's predictions more often than the other's. The percentage matching each model's predictions ranged from 42.1\% to 57.9\%, with no significant departure from an equal split after FDR correction across the six pairs (all adjusted $p\geq.164$; Supplementary Fig.~\ref{fig:model_alignment_matrix}).

\FloatBarrier
\FloatBarrier

\subsection{Model--human misalignment persists with strong human consensus and high human confidence}
\begin{figure}[ht!]
    \centering
    \includegraphics[width=\linewidth]{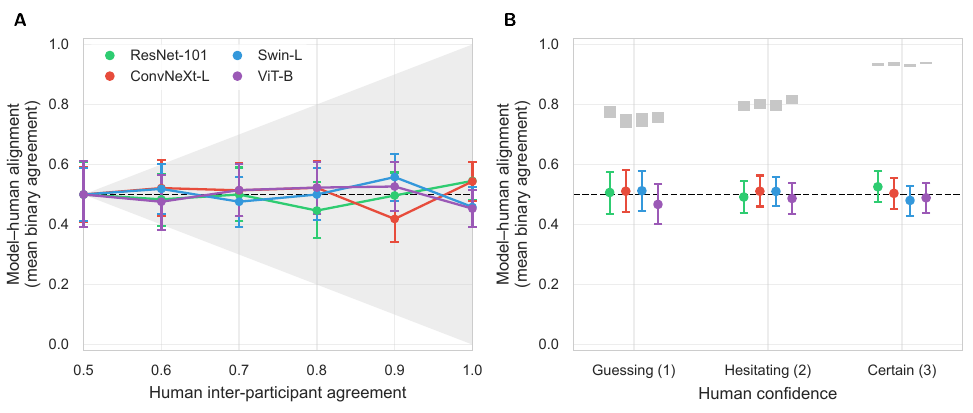}
    \caption{\textbf{Model--human alignment across levels of human consensus and human confidence.} Both panels evaluate each model only on stimuli generated to target it. \textbf{(A)} Mean binary model--human agreement at each level of inter-participant agreement---the proportion of the 10 participants giving the most common category response. A value of 0.5 indicates an equal split, and 1.0 indicates unanimity. The gray region shows the attainable range of mean binary agreement at each level of inter-participant agreement. Even for stimuli classified unanimously by participants, mean binary model--human agreement remained near chance. \textbf{(B)} Mean binary model--human agreement for responses grouped by reported confidence. Gray bars span the lower and upper bounds on the human noise ceiling for each group of responses. These bounds were estimated using responses from participants who viewed the same stimuli, regardless of those participants' reported confidence. Model--human agreement remained near chance even for responses rated \emph{certain}. Colored markers and error bars show means and 95\% two-factor bootstrap confidence intervals. Dashed lines mark chance level (0.5).}
    \label{fig:confidence_alignment}
\end{figure}

Could overall near-chance model--human agreement conceal better alignment for stimuli on which humans agreed more consistently? We therefore examined model--human agreement as a function of inter-participant agreement, which ranged from 0.5 for an evenly divided response to 1.0 for a unanimous response (Fig.~\ref{fig:confidence_alignment}A). Model--human agreement did not increase monotonically across this range. Even for stimuli classified unanimously by participants, mean model--human agreement ranged from 45.3\% to 54.5\% across the four models, and every corresponding confidence interval included 50\%. Thus, near-chance alignment was not confined to stimuli that elicited disagreement among participants.

To assess alignment when participants were confident about their category judgments, we pooled trials separately within each reported confidence level: Guessing, Hesitating, and Certain (Fig.~\ref{fig:confidence_alignment}B). Mean binary agreement remained near chance at all three levels; all 12 model-specific 95\% confidence intervals included 50\%. On trials rated Certain, model means ranged from 48.0\% to 52.6\%, whereas the corresponding lower human noise-ceiling bounds ranged from 92.6\% to 93.5\%. Thus, model--human agreement remained near chance even for participants' most confident judgments.

\FloatBarrier

\subsection{Confidence-sensitive scoring reveals limited model--human correspondence}
So far, the analyses have reduced both human responses and model outputs to binary category judgments. We next used signed-rank cosine similarity (SRCS; Section~\ref{sec:quantifying-alignment}) to take human and model confidence levels into account. SRCS exceeded chance level for the two CNNs, ResNet-101 ($\mathrm{SRCS}=.102$, adjusted $p=.018$) and ConvNeXt-L ($\mathrm{SRCS}=.148$, adjusted $p=.0008$). SRCS did not differ significantly from chance for Swin-L ($\mathrm{SRCS}=.074$, adjusted $p=.070$) or ViT-B ($\mathrm{SRCS}=.021$, adjusted $p=.625$; Fig.~\ref{fig:model_comparison}B). Mean SRCS did not differ significantly between models in any of the six pairwise comparisons (all adjusted $p\geq.239$). All four SRCS estimates were below their corresponding lower human noise-ceiling bounds (all adjusted $p=.0002$). Overall, the confidence-sensitive analysis provided limited evidence of weak model--human alignment for the two CNNs.

\subsection{Misalignment is strongest for models targeted by the optimization}
In the analyses above, we evaluated each model only on stimuli generated to target it. This provided a stringent test of model--human alignment on images optimized to expose disagreements with another model. However, models not targeted during a stimulus's optimization can also make a prediction about its category. We asked whether predictions from these \emph{non-targeted} models agreed more closely with human judgments.

\begin{figure}[ht!]
    \centering
    \renewcommand{\thesubfigure}{\Alph{subfigure}}
    \begin{subfigure}[t]{0.5\linewidth}
        \centering
        \makebox[\linewidth][l]{\sffamily\bfseries\small A}
        \includegraphics[width=\linewidth]{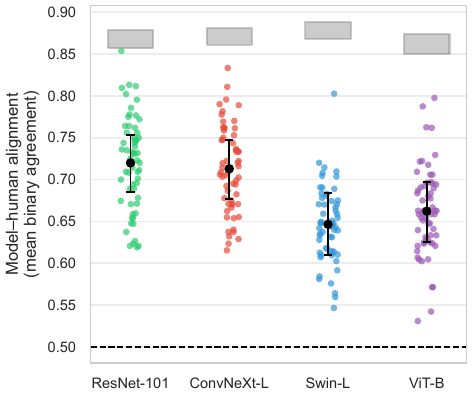}
        \phantomsubcaption\label{fig:non_participating_accuracy}
    \end{subfigure}
    \hfill
    \begin{subfigure}[t]{0.49\linewidth}
        \centering
        \makebox[\linewidth][l]{\sffamily\bfseries\small B}
        \includegraphics[width=\linewidth]{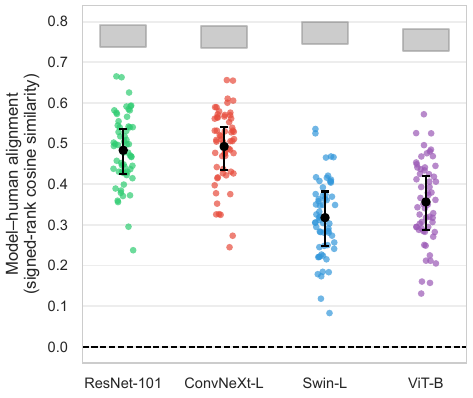}
        \phantomsubcaption\label{fig:signed-non-participating}
    \end{subfigure}
    \caption{\textbf{Model--human alignment for the four models on non-targeted stimuli.} For both measures, each model is evaluated only on stimuli generated using pairs of the other three models. \textbf{(A)} Mean binary model--human agreement. \textbf{(B)} Signed-rank cosine similarity (SRCS). Colored points show participant-level values; black markers and error bars show group means and 95\% two-factor bootstrap confidence intervals. Dashed lines mark chance-level performance. Gray bars span the lower and upper bounds on the human noise ceiling.}
    \label{fig:primary_non_targeted_alignment}
\end{figure}

On non-targeted stimuli, mean binary model--human agreement ranged from 64.6\% to 72.0\% (Fig.~\ref{fig:primary_non_targeted_alignment}A). All four models exceeded chance (all adjusted $p=.0002$), but their mean binary agreement remained below the corresponding lower human noise-ceiling bounds (all adjusted $p=.0002$). Mean binary agreement differed significantly between models in three of the six pairwise comparisons after FDR correction: ResNet-101 had higher agreement than ViT-B (adjusted $p=.041$) and Swin-L (adjusted $p=.014$), and ConvNeXt-L had higher agreement than Swin-L (adjusted $p=.014$).

For all four models, SRCS also exceeded chance level (all adjusted $p=.0002$), but remained below the corresponding lower human noise-ceiling bounds (all adjusted $p=.0002$; Fig.~\ref{fig:primary_non_targeted_alignment}B). Four of the six pairwise SRCS comparisons were significant after FDR correction: ResNet-101 and ConvNeXt-L had higher SRCS than both ViT-B and Swin-L (all adjusted $p\leq.0039$). Near-chance binary agreement was therefore specific to models targeted during stimulus generation.

\FloatBarrier

\subsection{Model--human alignment across 17 additional robust models}
To assess alignment beyond the four models used for stimulus generation, we evaluated 17 additional top-ranked adversarially robust ImageNet classifiers from RobustBench \citep{croce2021robustbench} (Supplementary Table~\ref{tab:additional-models}). All 17 models were evaluated on all 1,008 stimuli; none was used to generate or select the stimuli.

\begin{figure}[ht!]
    \centering
    \renewcommand{\thesubfigure}{\Alph{subfigure}}
    \begin{subfigure}[t]{\linewidth}
        \centering
        \makebox[\linewidth][l]{\sffamily\bfseries\small A}
        \includegraphics[width=\linewidth]{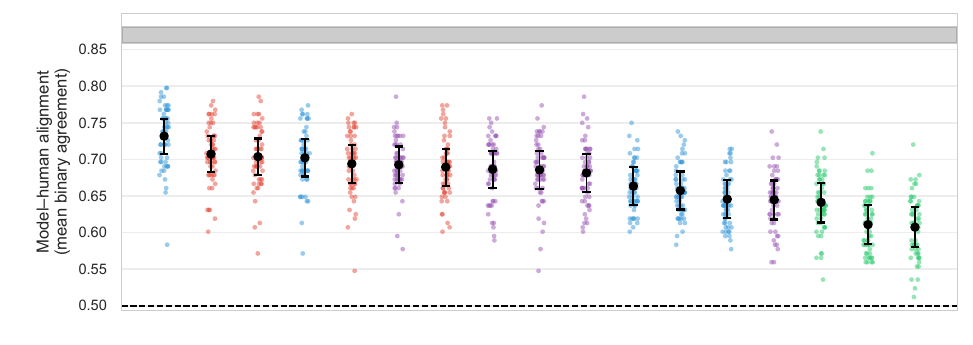}
        \phantomsubcaption\label{fig:additional_zoo_accuracy}
    \end{subfigure}
    \par\medskip
    \begin{subfigure}[t]{\linewidth}
        \centering
        \makebox[\linewidth][l]{\sffamily\bfseries\small B}
        \includegraphics[width=\linewidth]{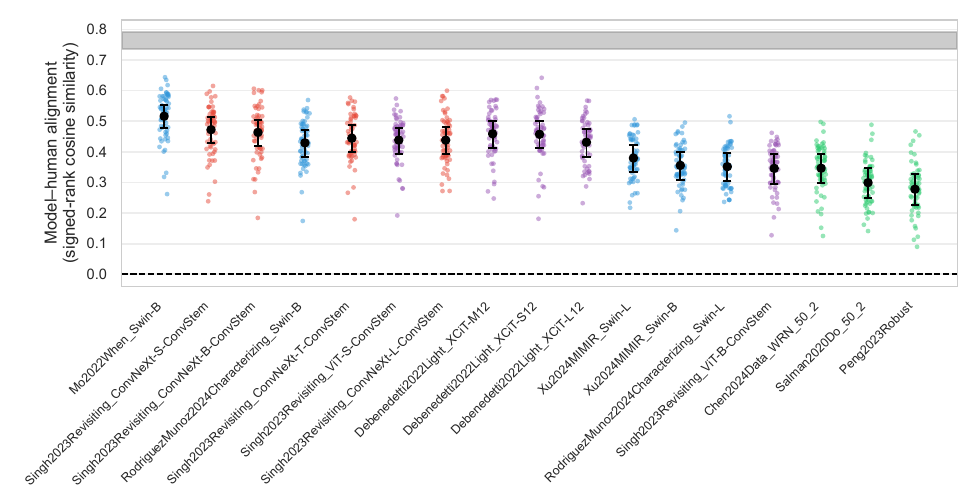}
        \phantomsubcaption\label{fig:signed-zoo}
    \end{subfigure}
    \caption{\textbf{Model--human alignment for the 17 additional robust models.} All 17 models were evaluated on all 1,008 stimuli; none was used to generate or select the stimuli. \textbf{(A)} Mean binary model--human agreement. \textbf{(B)} Signed-rank cosine similarity (SRCS). Colored points show participant-level values; black markers and error bars show group means and 95\% two-factor bootstrap confidence intervals. Dashed lines mark chance-level performance. In each panel, the gray band spans the lower and upper bounds on the human noise ceiling shared by all 17 models. Colors identify Swin (blue), ConvNeXt (red), ViT/XCiT (purple), and CNN/WRN (green) models. Models are ordered from left to right by decreasing mean binary model--human agreement; the model identifiers in B apply to both panels.}
    \label{fig:generalization}
\end{figure}

Mean binary model--human agreement ranged from 60.7\% to 73.2\% (Fig.~\ref{fig:generalization}A), and SRCS ranged from .278 to .516 (Fig.~\ref{fig:generalization}B). The lower and upper human noise-ceiling bounds (shared across all 17 models) were 85.9\% and 88.1\% for binary agreement and .736 and .791 for SRCS. All 17 models significantly exceeded chance on both measures (all adjusted $p=.0002$), but agreement remained significantly below the corresponding lower noise-ceiling bounds (all adjusted $p=.0002$). The gap to human reliability therefore persisted across this broader set of top-ranked robust models.

\FloatBarrier

We asked whether the variation in model--human agreement was associated with parameter count, clean accuracy, or PGD100 accuracy (Supplementary Fig.~\ref{fig:model_property_correlations}). The six estimated Spearman correlations ranged from $\rho=-.312$ to $\rho=.147$, and none was significant after FDR correction across these six correlations (all adjusted $p\geq.970$). These results do not suggest that more adversarially robust models are better aligned with human judgments.

\subsection{Model--human disagreement exceeds inter-observer disagreement}

A model could disagree with the group average while still closely matching the judgments of an individual observer. We therefore tested whether the 17 additional models agreed with individual participants as closely as participants agreed with one another (Fig.~\ref{fig:individual_observer_agreement}).

Participants agreed substantially more with one another than the models agreed with them. Mean participant--participant agreement was 81.3\%, compared with 67.3\% mean model--participant agreement, a difference of 14.0 percentage points (95\% two-factor bootstrap CI [11.9, 16.0]). Mean agreement between pairs of models was similarly low, at 67.2\%. Thus, even for models not targeted during image optimization, model--human discrepancy exceeded the disagreement among individual observers on the same images.

\section{Discussion}
\label{sec:discussion}
We found marked disagreement between four adversarially trained ImageNet classifiers and human category judgments on GAN-constrained controversial stimuli. When directly targeted by the optimization, each model showed approximately chance-level agreement with human category choices, despite substantial agreement among observers. This disparity persisted for stimuli classified unanimously and for human choices made with the highest reported confidence. Comparing graded model predictions with graded human judgments revealed limited correspondence for two models, but all four remained below their human noise ceilings. Agreement with humans was higher when the same images were evaluated by classifiers not directly targeted during image optimization, including 17 additional robust classifiers, although all remained below the corresponding lower noise-ceiling bounds.

\subsection{Model-specific and more general misalignment}

The near-chance model--human agreement of all four targeted models (Fig.~\ref{fig:model_comparison}A) was not forced by the construction of the stimuli. Had one of these models correctly specified the average observer's category-choice probabilities, it could have achieved high model--human agreement even on stimuli targeting it. We simulated this scenario by deriving human category-choice probabilities from each model's predictions in turn, calibrating their variability to reproduce the empirical human noise-ceiling bounds, and sampling human judgments from the resulting probabilities. For each simulated dataset, the model whose predictions defined the human category-choice probabilities achieved 85.5--87.9\% model--human agreement (Supplementary Fig.~\ref{fig:synthetic_targeted_alignment}).

The higher model--human agreement of non-targeted models suggests that the targeted discrepancies contained a substantial model-specific component. This pattern is consistent with the model-specific invariances reported by \citet{feather2023model}: human recognition of a model's metamers closely tracked their recognition by other models. Evaluating a classifier only on stimuli optimized for other models may therefore miss discrepancies revealed when the classifier itself is targeted.

\citet{kriegeskorte2015deep} suggested that biological visual systems may likewise have idiosyncratic adversarial vulnerabilities, which may remain concealed because directly targeting an individual observer is experimentally difficult. We therefore cannot determine whether the discrepancies exposed by directly targeting a model have analogous observer-specific counterparts in humans. However, the comparison of non-targeted models with individual observers shows that current robust classifiers would be highly atypical if treated as draws from the same population as the human participants: the 17 additional models agreed less with participants than participants agreed with one another on the same images (Supplementary Fig.~\ref{fig:individual_observer_agreement}).

The observed model--human misalignment therefore indicates both pronounced model-specific discrepancies exposed by targeted optimization and a broader model--human gap that persists when the evaluated model was not targeted.

\subsection{Local robustness and perceptual category boundaries}

At first sight, the present findings may seem difficult to reconcile with evidence that adversarial training can improve correspondence with biological vision \citep{tsipras2019robustness,feather2023model,gaziv2023strong,gaziv2025noninvasive,prince2026parametric,croce2025perceptual} (see also recent evidence for human audition, obtained using stimuli controversial between robust and non-robust models \citep{skrill2026neural}). Such findings show that robust models can share important sensitivities and invariances with biological systems \citep{tong2026ecological}. Our results show that this partial correspondence can coexist with substantial model--human disagreement over category membership.

The present findings do not indicate a failure of robustness under the norm-bounded threat models used to adversarially train these models. The observed model--human disagreement highlights the limited scope of this technical notion of robustness: it constrains model predictions only within prescribed norm-bounded neighborhoods of natural images. Beyond these neighborhoods, the adversarial objective provides no direct guidance for where category boundaries should lie. Our controversial stimuli were not restricted to such neighborhoods, yet human observers often categorized them consistently, revealing substantial model--human disagreement in regions that the robustness objective does not directly constrain.

Consistent with this distinction, conventional robustness benchmark performance did not track model--human alignment across the 17 held-out robust models. Larger models tended to achieve higher clean and adversarial accuracy, but this was not accompanied by higher human alignment. We therefore infer that further gains in $\ell_\infty$ robustness on RobustBench should not, by themselves, be expected to improve human alignment as measured here.

One possible direction is to use classification objectives that model the class-conditional image distribution rather than only the mapping from images to labels. On MNIST and CIFAR-10, generative and hybrid generative--discriminative classifiers predicted human category judgments better than the purely discriminative models tested \citep{golan2020controversial}. More recently, text-to-image generative classifiers showed human-like shape bias, near-human out-of-distribution accuracy, and strong correspondence with human error patterns \citep{jaini2024intriguing}. These findings suggest that generative objectives may constrain classification in regions where a locally robust discriminative objective supplies no direct training signal. Whether generative classifiers yield more human-aligned category boundaries at ImageNet scale remains to be tested.

\subsection{Limitations}
Our analysis was limited to models developed for the ImageNet threat model evaluated in RobustBench---$\ell_\infty$ perturbations with $\epsilon=4/255$. Whether the results generalize to other definitions of robustness remains open.

Because we used pretrained models, modeling choices were not experimentally controlled. The four targeted models were relatively comparable in training setup. The 17 additional RobustBench models were more heterogeneous, differing in adversarial training method, pretraining, and data filtering and augmentation. We therefore do not make causal claims about differences in human alignment between models, and cannot attribute such differences specifically to architecture. 

Finally, because only a single checkpoint was available for each pretrained model, we cannot distinguish variation associated with a model family from variation specific to a particular trained instance (e.g., effects of weight initialization).

\subsection{Conclusion}

Adversarial robustness does not entail human-aligned categorization. A classifier can satisfy a prescribed robustness criterion and thus share some invariances with biological visual systems, yet retain category boundaries that differ substantially from those of humans. Robustness therefore does not preclude marked mismatches between model categorizations and human judgments; establishing human alignment requires direct behavioral evaluation rather than inference from performance on in silico robustness benchmarks.

\subsection*{Declaration of generative AI and AI-assisted technologies in the manuscript preparation process}
During the preparation of this work, the authors used ChatGPT and Codex (OpenAI) and Claude and Claude Code (Anthropic) to assist with code development and copyediting. The authors reviewed and verified all AI-assisted outputs and take full responsibility for the article's content.

\subsection*{Data and code availability}
All anonymized behavioral response data, experimental stimuli, model outputs, and code used for controversial-stimulus optimization, statistical analyses, and figure generation are available at:
\par\smallskip\noindent
\href{https://github.com/brainsandmachines/adversarial-robustness-does-not-entail-human-aligned-categorization}{\texttt{https://github.com/brainsandmachines/}}\\
\href{https://github.com/brainsandmachines/adversarial-robustness-does-not-entail-human-aligned-categorization}{\texttt{adversarial-robustness-does-not-entail-human-aligned-categorization}}.

\subsection*{CRediT authorship contribution statement}
Itay Reznik: Formal analysis, Methodology, Software, Visualization, Writing -- original draft. Raz Lapid: Conceptualization, Methodology, Software, Writing -- review and editing. Moshe Sipper: Conceptualization, Methodology, Supervision, Writing -- review and editing. Tal Golan: Conceptualization, Funding acquisition, Methodology, Supervision, Writing -- review and editing.

\subsection*{Funding}
This research was supported by the Israel Science Foundation (grant No. 534/24). The funder had no role in the study design; the collection, analysis, or interpretation of the data; the writing of the manuscript; or the decision to submit the article for publication.

\subsection*{Acknowledgments}
We thank Ohad Kelner for his assistance in programming the PsychoPy experiment and conducting the online testing.

\bibliographystyle{cas-model2-names}
\bibliography{refs}

\clearpage
\renewcommand{\thetable}{S\arabic{table}}
\setcounter{table}{0}
\renewcommand{\thefigure}{S\arabic{figure}}
\setcounter{figure}{0}
\section*{Supplementary Material}
\subsection*{Supplementary Tables}

\begin{minipage}{\textwidth}
\centering
\captionsetup{hypcap=false}
\captionof{table}{Seventeen additional RobustBench ImageNet classifiers. Parameters are reported in millions. Clean accuracy is top-1 accuracy on unperturbed images. PGD100 and AutoAttack are top-1 accuracies under an $\ell_\infty$ perturbation budget of $\epsilon=4/255$. Accuracies were measured on the same ARES sample of 1,000 ImageNet validation images used for Table~\ref{tab:robust-models}. Rows are ordered by increasing PGD100 accuracy. Models built on the checkpoints of generating models (\texttt{Amini2024MeanSparse\_\allowbreak ConvNeXt-L}, \texttt{Amini2024MeanSparse\_\allowbreak Swin-L}, and \texttt{Bai2024MixedNUTS}) were excluded.}
\label{tab:additional-models}
\footnotesize
\setlength{\tabcolsep}{3pt}
\renewcommand{\arraystretch}{1.05}
\begin{tabular}{p{0.40\linewidth} r r r r}
\hline
RobustBench model identifier & Parameters (M) & Clean acc. (\%) & PGD100 acc. (\%) & AutoAttack acc. (\%) \tabularnewline
\hline
\texttt{Mo2022When\_\allowbreak Swin-B} & 87.8 & 75.3 & 42.3 & 38.8 \tabularnewline
\texttt{Salman2020Do\_\allowbreak 50\_\allowbreak 2} & 68.9 & 70.5 & 43.7 & 40.0 \tabularnewline
\texttt{Debenedetti2022Light\_\allowbreak XCiT-S12} & 26.3 & 74.5 & 46.0 & 43.1 \tabularnewline
\texttt{Chen2024Data\_\allowbreak WRN\_\allowbreak 50\_\allowbreak 2} & 68.9 & 69.6 & 48.4 & 42.4 \tabularnewline
\texttt{Debenedetti2022Light\_\allowbreak XCiT-M12} & 46.4 & 75.3 & 50.1 & 47.0 \tabularnewline
\texttt{Singh2023Revisiting\_\allowbreak ViT-S-ConvStem} & 22.8 & 73.4 & 51.8 & 49.5 \tabularnewline
\texttt{Peng2023Robust} & 104.1 & 74.6 & 51.9 & 49.6 \tabularnewline
\texttt{Debenedetti2022Light\_\allowbreak XCiT-L12} & 103.8 & 75.8 & 52.1 & 49.6 \tabularnewline
\texttt{Singh2023Revisiting\_\allowbreak ConvNeXt-T-ConvStem} & 28.6 & 73.4 & 55.1 & 50.1 \tabularnewline
\texttt{RodriguezMunoz2024Characterizing\_\allowbreak Swin-B} & 87.8 & 79.1 & 55.6 & 53.1 \tabularnewline
\texttt{Singh2023Revisiting\_\allowbreak ConvNeXt-S-ConvStem} & 50.3 & 75.0 & 56.1 & 53.6 \tabularnewline
\texttt{Singh2023Revisiting\_\allowbreak ViT-B-ConvStem} & 87.1 & 76.8 & 57.9 & 56.0 \tabularnewline
\texttt{RodriguezMunoz2024Characterizing\_\allowbreak Swin-L} & 196.5 & 79.6 & 58.9 & 55.5 \tabularnewline
\texttt{Xu2024MIMIR\_\allowbreak Swin-B} & 87.8 & 77.6 & 59.9 & 57.7 \tabularnewline
\texttt{Singh2023Revisiting\_\allowbreak ConvNeXt-B-ConvStem} & 88.8 & 77.1 & 61.0 & 58.5 \tabularnewline
\texttt{Singh2023Revisiting\_\allowbreak ConvNeXt-L-ConvStem} & 198.1 & 78.1 & 61.6 & 60.5 \tabularnewline
\texttt{Xu2024MIMIR\_\allowbreak Swin-L} & 196.8 & 79.0 & 64.4 & 61.4 \tabularnewline
\hline
\end{tabular}
\end{minipage}

\newpage

\subsection*{Supplementary Figures}

\begin{figure}[ht!]
    \centering
    \includegraphics[width=0.7\linewidth]{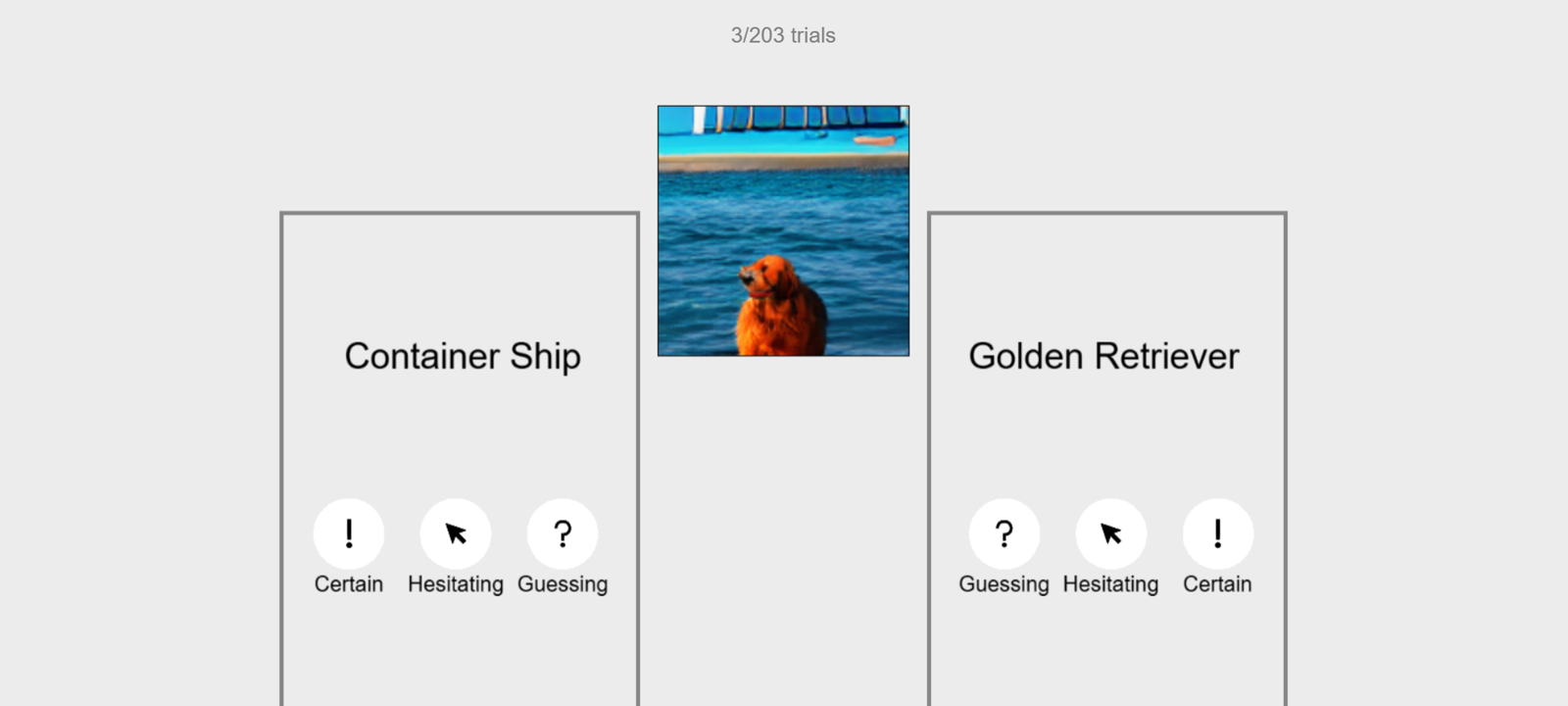}
    \caption{\textbf{Example trial from the behavioral experiment.} An image was presented centrally with the two candidate category labels. Each label was paired with three confidence options---\emph{certain}, \emph{hesitating}, and \emph{guessing}. Participants selected one of the six resulting response buttons to report both their category judgment and confidence. The left-right positions of the two labels were randomized across trials. Participants were unaware that the two category labels represented opposing predictions from two models.}
    \label{fig:behavioral-trial}
\end{figure}

\begin{figure}[ht!]
    \centering
    \includegraphics[width=\linewidth]{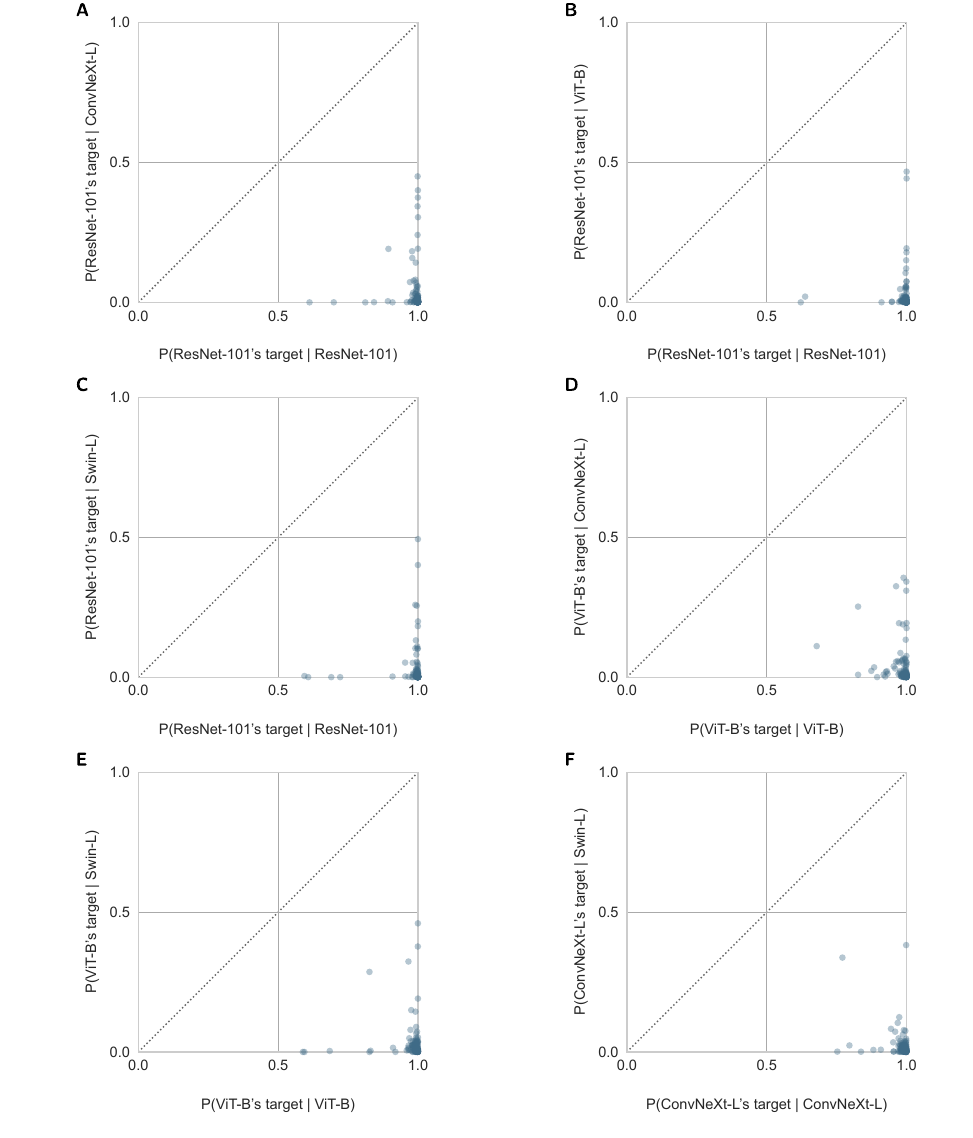}
    \caption{\textbf{Opposing category preferences for all 1,008 experimental stimuli.} Each panel plots the probabilities assigned by two models to the same target category for the 168 experimental stimuli generated for that model pair. Each dot represents one stimulus. The category shown on both axes is the horizontal-axis model's target category. $P(\text{category}\mid\text{model})$ denotes the probability assigned to that category by the model, normalized over the two candidate categories. The dotted diagonal marks equal probabilities between models, and reference lines at 0.5 mark equal probabilities for the two categories. \textbf{Dots near the bottom-right corner indicate strong disagreement: each model assigned high normalized probability to its own target category.}}
    \label{fig:pairwise-target-probabilities}
\end{figure}

\begin{figure}[ht!]
    \centering
    \includegraphics[width=\linewidth]{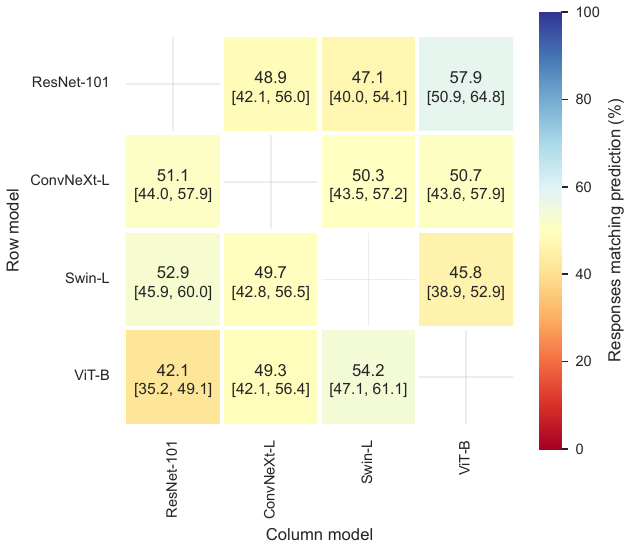}
    \caption{\textbf{Human agreement with opposing model predictions.} Each off-diagonal cell shows the percentage of human responses matching the row model's prediction on stimuli generated using the row and column models. The two percentages for each model pair sum to 100\%. Bracketed values give 95\% two-factor bootstrap confidence intervals. Responses did not differ significantly from an equal split for any of the six model pairs after FDR correction.}
    \label{fig:model_alignment_matrix}
\end{figure}

\begin{figure}[p]
    \centering
    \includegraphics[width=\linewidth]{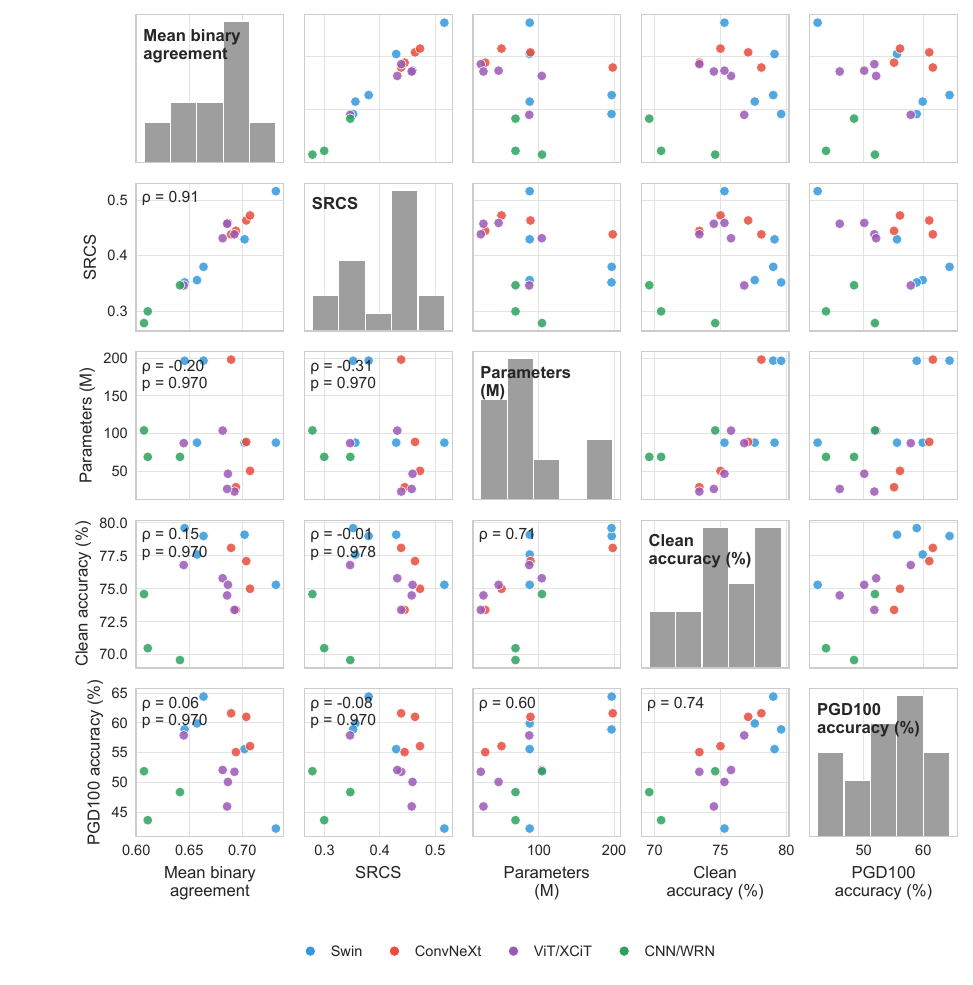}
    \caption{\textbf{Relationships among model--human agreement, parameter count, and clean and adversarial classification accuracy.} Model--human agreement was quantified using mean binary agreement and signed-rank cosine similarity (SRCS). Clean and PGD100 accuracies were computed on the 1,000-image ARES validation sample (Supplementary Table~\ref{tab:additional-models}). Each point represents one of the 17 additional models. Diagonal panels show histograms; off-diagonal panels show pairwise associations. Lower-triangle annotations report Spearman's $\rho$. Two-sided $p$-values are shown for the six correlations between agreement measures and model properties, with FDR correction across these tests. AutoAttack accuracy was omitted because its rank correlation with PGD100 accuracy was near-perfect (Spearman $\rho=.994$).}
    \label{fig:model_property_correlations}
\end{figure}

\begin{figure}[p]
    \centering
    \includegraphics[width=\linewidth]{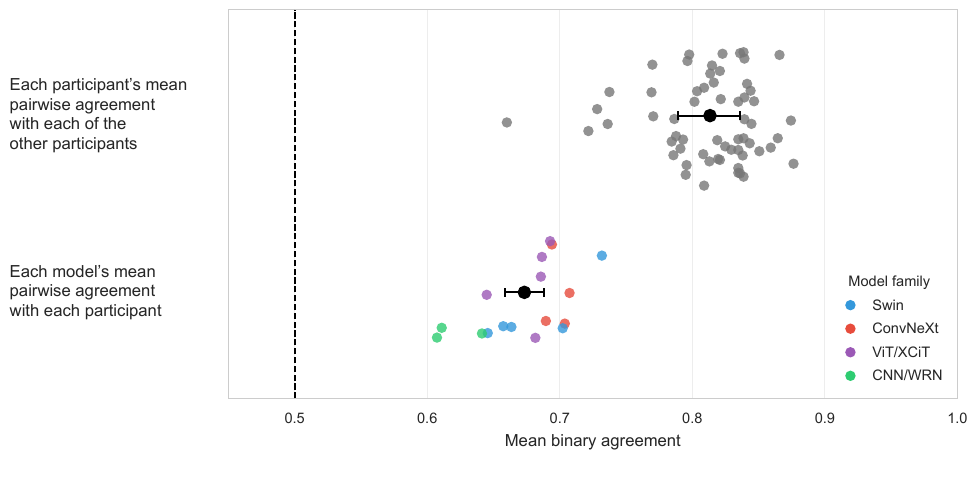}
    \caption{\textbf{Mean human--human agreement exceeded model--human agreement.} At the top, gray points show the 60 participants' human--human agreement scores. Each score is the mean, across that participant's 168 images, of the fraction of the other nine raters who made the same category choice. Below, colored points show the 17 additional models' mean binary agreement with humans, calculated within each participant's 168-image set and then averaged across all 60 participants. All models were evaluated on the same 1,008 images and none participated in stimulus generation or selection. Model choices were determined by the larger logit for the two candidate categories. Black markers and horizontal error bars show group means and 95\% participant-and-stimulus bootstrap confidence intervals, with the 17 models held fixed. Colors identify model families. Points are spread vertically for visibility. The dashed line marks 50\% agreement. Mean human--human and model--human agreement were 81.3\% and 67.3\%, respectively (difference: 14.0 percentage points; 95\% confidence interval: 11.9--16.0).}
    \label{fig:individual_observer_agreement}
\end{figure}

\begin{figure}[p]
    \centering
    \includegraphics[width=\linewidth]{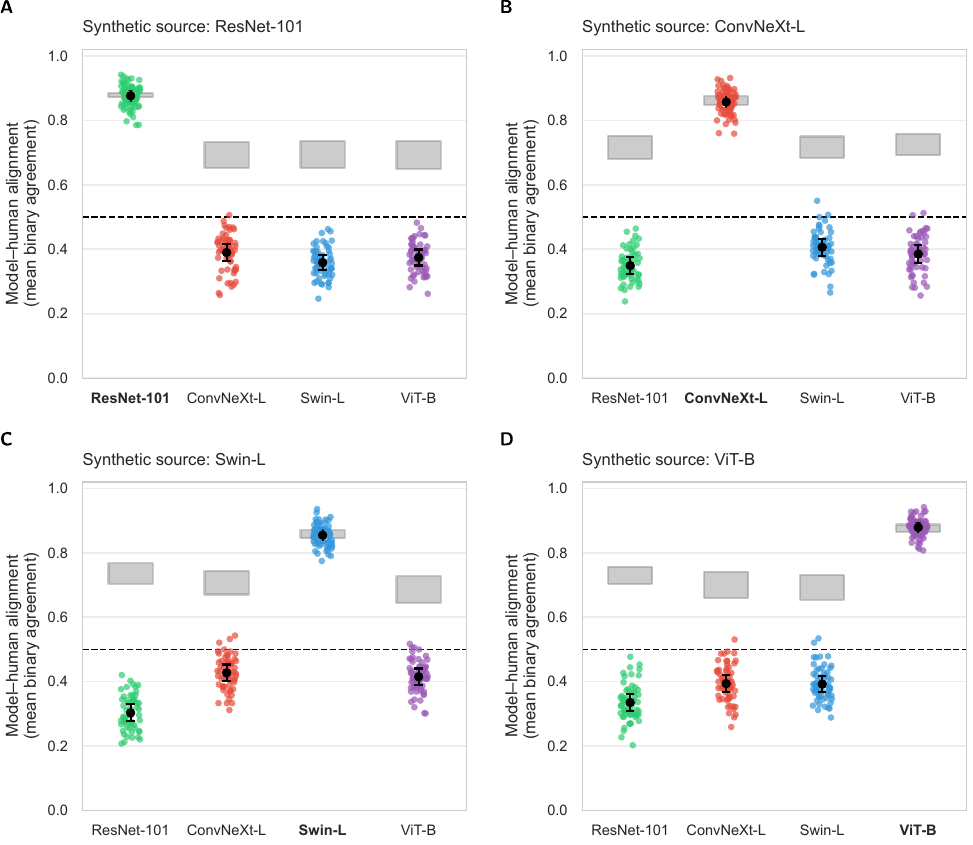}
    \caption{\textbf{A targeted classifier could achieve high model--human agreement if it captured the average human observer's category preferences.} We simulated four experimental datasets, each with 60 participants whose category-choice probabilities were determined by ResNet-101 (A), ConvNeXt-L (B), Swin-L (C), or ViT-B/16 (D). Each dataset preserved the original participant--image assignments: 168 images per participant and ten responses per image across 1,008 images. Responses were sampled independently from probabilities that preserved the model's category preferences, with variability adjusted to match the empirical noise-ceiling bounds on the 504 images targeting that model. We repeated the targeted-model agreement analysis for all four classifiers on each dataset, evaluating each classifier on its own 504 targeted images. Colored points show participant-level binary agreement; black markers and error bars show means and 95\% participant-and-stimulus bootstrap intervals. Gray bars span the lower and upper synthetic noise ceilings for each classifier's targeted images. The bold label identifies the model determining the simulated responses; dashed lines mark 50\% agreement. The classifier determining the simulated responses achieved 85.47--87.88\% agreement on its own targeted images, whereas all four classifiers showed near-chance agreement with human responses in the experiment (Fig.~\ref{fig:model_comparison}A).}
    \label{fig:synthetic_targeted_alignment}
\end{figure}

\FloatBarrier
\end{document}